\documentclass[letterpaper, 10 pt, conference]{ieeeconf}  

\IEEEoverridecommandlockouts                              

\usepackage{xcolor}
\usepackage{amsmath} 
\usepackage{amssymb}  
\usepackage{graphicx}
\usepackage[ruled,vlined]{algorithm2e}
\usepackage{microtype}
\usepackage{relsize}

\title{\LARGE \bf
Online Evolution Strategy for Flow-Matching VLA Policies via Self-Supervised Trajectory Distribution Optimization
}

\author{Gongxin Yao, Yongsheng Zhao, Jiayin Deng, Deng Liang, Han Gao, Lei Zhao, Baoping Cheng*
\thanks{These authors are with the China Mobile (Hangzhou) Information Technology Co., Ltd., Hangzhou, 311121, China. (Baoping Cheng* is the corresponding author, email: chengbaoping@cmhi.chinamobile.com). }%
}

\begin{document}

\maketitle
\thispagestyle{empty}
\pagestyle{empty}

\begin{abstract}
Vision-Language-Action (VLA) models based on generative frameworks, such as Flow Matching, have recently achieved impressive performance in robotic manipulation. Unlike deterministic policies, Flow Matching enables VLA models to learn conditional action trajectory distributions, where latent noise vectors induce different actions under the same task scenario. However, we observe that these distributions are often ill-formed, with successful and failed behaviors coexisting while considerable probability mass remains in unfavorable regions. To this end, we propose Online-ES, an online adaptation framework for Flow Matching VLAs based on Evolution Strategy (ES), which refines the learned action trajectory distribution through interaction feedback. Instead of pruning the latent noise space, our method performs evolutionary exploration directly in the action trajectory space, where diverse trajectories generated by Flow Matching provide candidate solutions for adaptation. By perturbing sampled trajectories and evaluating their execution outcomes, we derive a self-supervised MSE objective that transfers the evolution direction from trajectory space into model parameter space. Mathematically, we prove that the proposed objective provides an unbiased estimator of the optimal evolution direction. Moreover, we also incorporate failure experiences as negative feedback to regularize the evolution direction, steering the policy away from previously explored failure regions. Experiments in both simulation and real-world environments demonstrate that Online-ES achieves policy improvement comparable to reinforcement fine-tuning, without learning a value model or computing advantages.

\end{abstract}

\section{INTRODUCTION}

Vision-Language-Action (VLA) models \cite{ma2026survey} have recently emerged as a promising paradigm for general-purpose robot manipulation. Instead of learning semantic representations from scratch, a VLA model typically integrates a pretrained vision-language model \cite{zhang2024vision} (VLM) with an action expert, where the VLM provides powerful grounding and reasoning capabilities to interpret visual observations and task instructions. The action expert is then trained on human teleoperation \cite{si2021review} datasets to map high-level VLM representations into low-level robot actions. In earlier studies \cite{shridhar2022cliport, zitkovich2023rt, yin2023multi}, conventional neural network architectures, such as MLPs or Transformers, were commonly adopted as action experts to regress robot actions from observations. However, robot manipulation is inherently multimodal \cite{zhai2025vfp}, where multiple feasible action trajectories may achieve the same task objective within the same scenario. Consequently, these action experts tend to produce a single deterministic trajectory, collapsing diverse solutions into averaged behaviors. 

Flow Matching is a generative framework \cite{lipman2023flowmatching} that learns a continuous velocity field to transport samples from noise distribution to the target data distribution through an ordinary differential equation (ODE). It has achieved remarkable success in image and video generation \cite{xiao2025omnigen, kodaira2026streamdit}, demonstrating the ability to produce high-quality and diverse content. Recently, an increasing number of VLA models \cite{pmlr-v305-black25a, bjorck2025gr00t, wu2026foundation} have adopted Flow Matching as a new paradigm for action generation. Unlike deterministic action prediction, Flow Matching VLA models implicitly learn an action trajectory distribution conditioned on observation, where random latent noise vectors serve as the initial states of action generation, enabling diverse trajectories to be sampled from the learned distribution. Although such intrinsic stochasticity provides behavioral diversity, it also introduces variability in execution results. Specifically, the generation process may occasionally sample an unfavorable latent noise vector, resulting in task failure. Meanwhile, other latent noise vectors may generate successful trajectories for the same task. This leads to our key insight: task failure in a single rollout may arise from the ill-formed action trajectory distribution, rather than the policy's inability to solve the task.

\begin{figure}[t]
    \centering
    \includegraphics[width=1\linewidth]{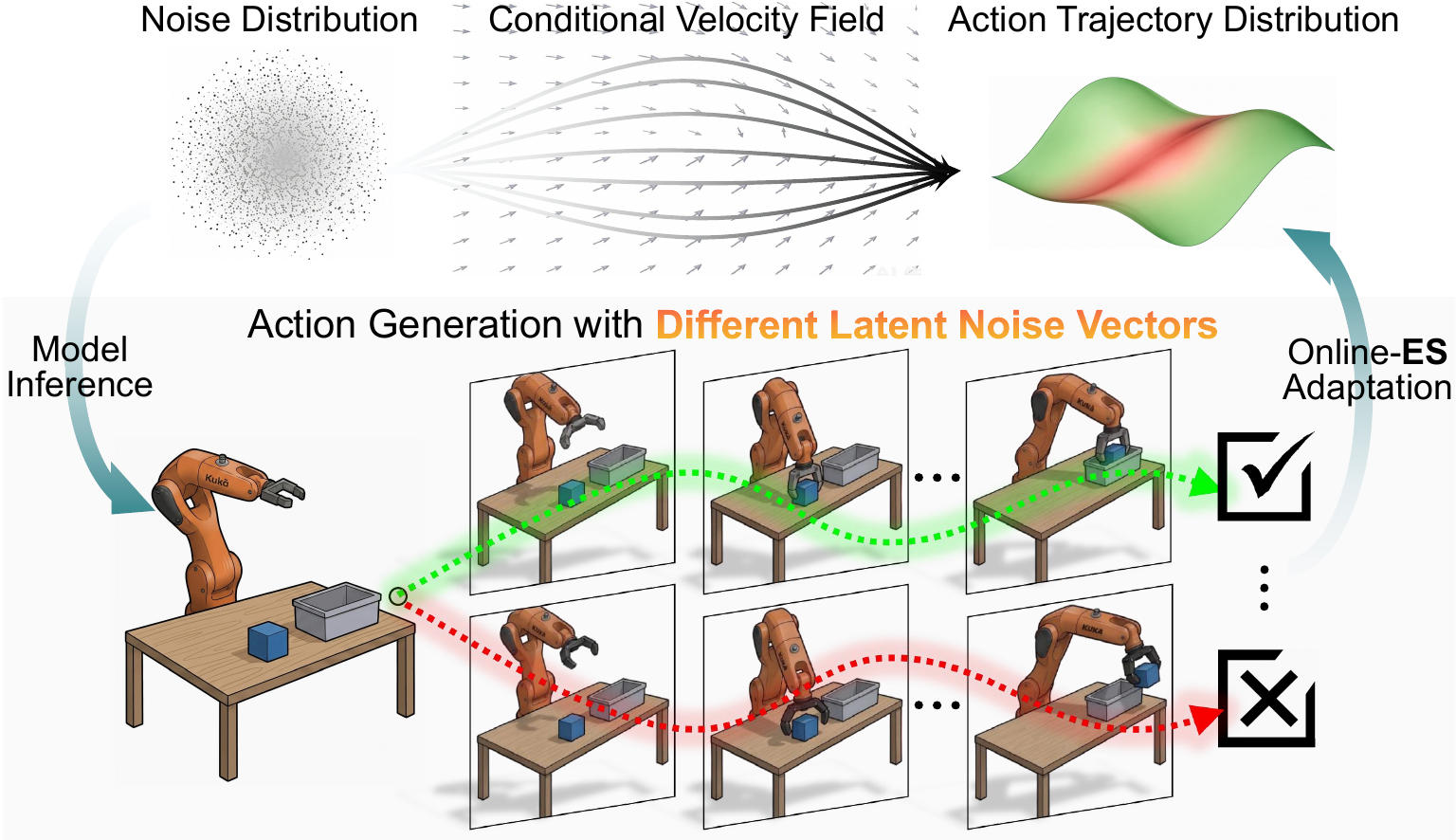}
    \vspace{-6mm}
    \caption{Flow Matching VLAs learn a conditional velocity field that maps noise distributions to action trajectory distributions. For each task scenario, we repeatedly sample different latent noise vectors to generate diverse action trajectories and execute them. Experiences from both successful and failed trajectories are then used to adapt the model, progressively reshaping the action trajectory distribution toward regions with more favorable behaviors.}
    \label{fig:motivation}
\end{figure}

Motivated by the above analysis, an intuitive idea is to prune the sampling space of the latent noise vector \cite{stoica2025contrastive}, thereby steering future generations away from unfavorable regions. However, a long-horizon action trajectory typically consists of multiple action chunks \cite{zhao2023learning}, each generated based on the observations resulting from its preceding chunks. Although the latent noise vector for each action chunk is sampled independently, their individual effects on the final execution result are difficult to disentangle due to the Markovian dependency. Moreover, noise-space pruning is inherently task-specific, as the unfavorable regions of the noise space vary substantially across task scenarios, making pruning rules difficult to generalize. Instead, we seek to adapt the model online according to the execution feedback as illustrated in Fig.~\ref{fig:motivation}, thereby continuously refining the action trajectory distribution.

In this paper, we propose Online-ES, a simple online adaptation framework for Flow Matching VLAs based on Evolution Strategy~\cite{wierstra2014natural}. This framework exploits the stochasticity of Flow Matching to generate diverse candidate trajectories for evolutionary exploration in the action trajectory space. Specifically, for each task scenario, we repeatedly reset the environment and sample different latent noise vectors to obtain a set of action trajectories, which provide a Monte Carlo approximation of the underlying action trajectory distribution. Each sampled action trajectory then undergoes local exploration through Gaussian perturbations and is evaluated through interaction with the environment. Based on the feedback-weighted perturbation responses, we derive a self-supervised MSE objective that transfers the evolution direction from trajectory space into parameter space for model adaptation. Mathematically, we prove that the proposed objective provides an unbiased estimator of the optimal evolution direction, thereby guiding the trajectory distribution toward regions with higher expected feedback. Furthermore, by incorporating complementary negative feedback from failed experiences, we introduce an implicit repulsive force that discourages the policy from revisiting previously explored failure regions. Experiments in simulation and real-world environments demonstrate that Online-ES can achieve policy improvement comparable to reinforcement fine-tuning, without learning a value model \cite{lu2025vla} or computing complex advantages \cite{chen2025pirl, li2025vla}. In summary, our main contributions are:
\begin{itemize}
\item An online adaptation framework for Flow Matching VLAs based on Evolution Strategy, which performs evolutionary exploration in action trajectory space to refine the ill-formed trajectory distribution.

\item A self-supervised paradigm to optimize Flow Matching VLAs, which learns from self-generated interaction trajectories without learning an additional value model or computing complex advantages.

\item A failure-aware regularization mechanism that incorporates failure experiences as complementary negative feedback, steering the evolution direction away from previously explored failure regions.
\end{itemize}

\section{Related Work}
\subsection{Flow Matching for Generative Modeling}

Flow Matching \cite{lipman2023flowmatching} has emerged as a prominent framework for efficient generative modeling. Unlike diffusion models \cite{croitoru2023diffusion}, which rely on multi-step reverse denoising to generate samples from noise distributions, Flow Matching learns a time-dependent vector field that transports samples along a continuous probability path through an ordinary differential equation (ODE). Note that the initial noise vector specifies the starting point of ODE integration, and different initializations lead to distinct transport paths under the learned vector field, resulting in diverse samples from the target distribution \cite{lipman2023flowmatching}. Thus, this formulation provides a unified and efficient framework for modeling complex multimodal distributions, enabling remarkable advances in text \cite{hu2024flow}, image \cite{xiao2025omnigen}, and video \cite{kodaira2026streamdit} generation.

\subsection{Vision-Language-Action Models}
VLA models have emerged as a new paradigm for robotic policy learning by integrating pretrained vision-language backbones with action experts \cite{ma2026survey}. In real-world manipulation, a single task often admits multiple feasible solutions, e.g., the same object can be grasped from different poses or approached through different paths. There is an increasing trend toward adopting generative modeling frameworks \cite{pmlrv270chisari25a, chi2025diffusion} to learn such multimodal behaviors. Notably, recent studies, including $\pi_{0.5}$ \cite{pmlr-v305-black25a}, GR00T \cite{bjorck2025gr00t}, and some World Action Models \cite{ye2026world, yuan2026fast}, have explored Flow Matching-based action generation by learning vector fields conditioned on visual observations and language instructions. Moreover, recent studies have introduced reinforcement learning to further improve VLA policies beyond supervised fine-tuning, including VLA-RL~\cite{lu2025vla} with process-level value learning, $\pi_{RL}$~\cite{chen2025pirl} with group-relative advantages, and VLA-RFT~\cite{li2025vla} with world-model-based verified rewards. In contrast, our method exploits the inherent action trajectory diversity of Flow Matching VLAs for evolutionary exploration, achieving similar policy improvement without explicit value modeling or complex advantage computation.

\subsection{Evolution Strategy}

Evolution Strategy (ES) is a class of gradient-free optimization methods inspired by biological evolution principles \cite{huning1976evolutionsstrategie}, and has been widely applied to continuous black-box optimization problems. Natural Evolution Strategies \cite{wierstra2014natural} formulate evolutionary optimization from a probabilistic perspective by optimizing the expected fitness of a search distribution, providing a principled way to estimate optimization directions through stochastic perturbations. It has also been extended to optimize neural networks by applying stochastic perturbations to the parameters, demonstrating that models can be trained without backpropagation \cite{salimans2017evolution}. Recent advances in ES have further improved its effectiveness for black-box optimization by developing adaptive sampling and distribution update strategies, which enhance exploration efficiency and convergence performance \cite{hansen2016cma, li2023multitask, nomura2025cma}. In this work, we will extend ES to the action trajectory space of Flow Matching VLA policy for online adaptation.

\section{Preliminaries}
\subsection{Problem Formulation}
\label{sec:problem}
Given a Flow-Matching VLA policy $\pi_{\theta}(\cdot)$ that learns a velocity field conditioned on the observation $\boldsymbol{o}$ and language instruction $\boldsymbol{l}$, an action chunk is generated by integrating the following ordinary differential equation over $t\in[0,1]$:
\begin{align}
\label{eq:ode}
\begin{aligned}
\frac{d\boldsymbol{A}_t}{dt}&=\pi_{\theta}(\boldsymbol{A}_t,t,\boldsymbol{o},\boldsymbol{l}),\\
\boldsymbol{A}_{t=0}&=\boldsymbol{z}, \quad \boldsymbol{z}\sim\mathcal{N}(0,I),
\end{aligned}
\end{align}
where the result at $t=1$ is the final action chunk $\boldsymbol{A}$. Notably, the latent noise vector $\boldsymbol{z}$ specifies the initial condition of the ODE integration. Different realizations can thus lead to different actions under the same observation-language input:
\begin{align}
\boldsymbol{z} \neq \boldsymbol{z}'
\quad\Longrightarrow\quad
\boldsymbol{A} \neq \boldsymbol{A}'.
\end{align}

\begin{figure}[t]
    \centering
    \includegraphics[width=1\linewidth]{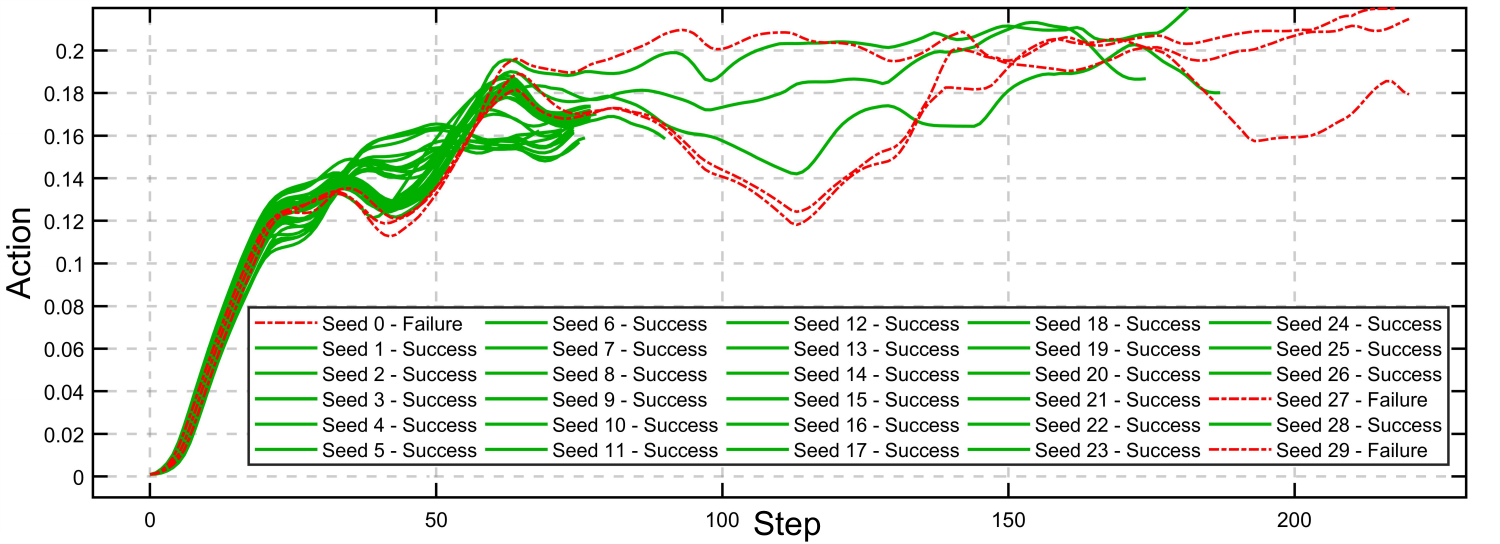}
    \vspace{-6mm}
    \caption{Action trajectories generated by a pretrained flow-matching VLA policy from different latent noise vectors within the same task scenario.}
    \label{fig:actiondist}
\end{figure}

In practice, the policy interacts with the environment to sequentially generate action chunks, with each chunk affecting the subsequent observations and action generations as:
\begin{align}
\boldsymbol{o}^{(0)}\rightarrow\boldsymbol{A}^{(0)}\rightarrow\boldsymbol{o}^{(1)}\rightarrow\boldsymbol{A}^{(1)}\rightarrow\boldsymbol{o}^{(2)}\rightarrow\boldsymbol{A}^{(2)}\rightarrow\cdots
\end{align}
where the superscripts denote the interaction steps and all chunks form a long-horizon action trajectory $\tau$. Although the variations induced by different latent noise vectors may be slight at early stages, this feedback loop allows the effects to accumulate over time, resulting in different trajectories over long horizons. For simplicity, we summarize this closed-loop generation process for a given task scenario $\mathcal{S}$ as:
\begin{align}
\label{eq:action_gen}
\tau=\pi_\theta(\mathcal{S},\boldsymbol{z}),
\quad
\boldsymbol{z}\sim\mathcal N(0,I),
\end{align}
which induces a conditional action trajectory distribution:
\begin{align}
\label{eq:vladist}
\tau \sim p_{\theta}(\tau|\mathcal{S}),
\end{align}
where each realization of $\boldsymbol{z}$ corresponds to a trajectory sampled from this distribution. Fig.~\ref{fig:actiondist} shows the results of a representative experiment: Under the same task scenario, different latent noise vectors produce diverse action trajectories, among which both successful and failed behaviors exist. This suggests that task failure in a single rollout may result from sampling an unfavorable latent noise vector, rather than the policy's inability to solve the task. Essentially, the coexistence of success and failure indicates that the action trajectory distribution in Eq.~\eqref{eq:vladist} is ill-formed, motivating us to refine it online.

\subsection{Evolution Strategy Fundamentals}
Evolution Strategy (ES) is a population-based optimization method inspired by natural selection, which iteratively explores the solution space through random perturbations and fitness-based selection. Given an objective function $J(\theta)$, ES generates a population of perturbed candidates around the current solution $\theta$ to explore its local neighborhood:
\begin{align}
    \label{eq:es_1}
    \theta_i = \theta + \epsilon_i\,, \; \epsilon_i \sim \mathcal{N}(0, \sigma^2I),
\end{align}
where $\sigma$ controls the perturbation scale. Then, the fitness of each candidate is computed as:
\begin{align}
    \label{eq:es_2}
    J(\theta_i)=J(\theta + \epsilon_i).
\end{align}
Here, $J(\theta)$ is defined as a maximization objective, where higher values indicate better solutions. The optimization direction is estimated by aggregating the perturbations weighted by their fitness as follows:
\begin{align}
    \label{eq:es_3}
    \nabla_{\theta} = \frac{1}{n \sigma^2}\sum^{n}_{i=1} J(\theta_i) \cdot \epsilon_i.
\end{align}
The solution is then updated as:
\begin{align}
    \label{eq:es_4}
    \theta \leftarrow \theta + \alpha \cdot \nabla_{\theta}, 
\end{align}
where $\alpha$ denotes the updating step size. By selecting perturbation directions associated with higher fitness, ES enables gradient-free optimization without explicit derivatives of the objective function. This strategy has been successfully extended to deep neural networks \cite{salimans2017evolution} as a scalable alternative to reinforcement learning.

\section{METHODOLOGY}
\subsection{Extending Evolution Strategy to Action Trajectory Space}
The goal of online adaptation is to refine the underlying trajectory distribution in Eq.~\eqref{eq:vladist}:
\begin{align}
\label{eq:target}
\max_{\theta}
\;\mathbb{E}_{\tau\sim p_{\theta}(\tau \mid \mathcal{S})}
\left[J(\tau)\right],
\end{align}
where $J(\tau)$ denotes the execution feedback from the environment. In robotic manipulation tasks, the feedback is typically formulated as a binary success indicator:
\begin{align}
\label{eq:reward}
J(\tau)=
\begin{cases}
1, & \text{if the task succeeds by executing } \tau,\\
0, & \text{otherwise},
\end{cases}
\end{align}
which makes Eq.~\eqref{eq:target} equivalent to maximizing the average success rate. 

Following prior work~\cite{salimans2017evolution}, a straightforward way to apply ES here is to perturb the policy parameters, thereby generating a population of candidate policies:
\begin{align}
\label{eq:es_space}
\pi_{\theta}(\cdot)\rightarrow \pi_{\theta+\epsilon}(\cdot), \;\; \epsilon \sim \mathcal{N}(0, \sigma^2I).
\end{align}
However, the massive number of parameters in VLA models makes such parameter-space exploration extremely inefficient. More critically, the deep architecture may amplify even small parameter perturbations, potentially leading to action trajectories that are physically infeasible or unsafe to execute. Therefore, we instead perform stochastic exploration directly in the action trajectory space as:
\begin{align}
\label{eq:es_dist}
p_{\theta}(\tau \mid \mathcal{S})
\rightarrow
p_{\theta}(\tau+\epsilon \mid \mathcal{S} ), \;\; \epsilon \sim \mathcal{N}(0, \sigma^2I).
\end{align}
By controlling the perturbation magnitude $\sigma$, we can preserve the base behavior of the pretrained VLA model while performing stochastic exploration in its local neighborhood. The beneficial explorations are then transferred to parameter space, thereby refining the action trajectory distribution.

\subsection{Evolution Direction of Action Trajectory Distribution}
\label{sec:evolution_direction}
Since an individual action trajectory only provides a limited view of the distribution implicitly represented by model parameters, we estimate its overall evolution direction through a sample-then-explore scheme. Specifically, given a resettable task scenario $\mathcal{S}$, we sample multiple latent noise vectors to generate a set of action trajectories as: 
\begin{align}
\label{eq:t_sample}
\Omega = \left\{ \tau_i \mid 
\tau_i=\pi_\theta(\mathcal{S},\boldsymbol{z}_i),
\boldsymbol{z}_i\sim\mathcal{N}(0,I)
\right\}_{i=1}^{N}.
\end{align}
It provides a Monte Carlo approximation of the underlying action trajectory distribution, as illustrated in Fig.~\ref{fig:actiondist}. For each action trajectory in $\Omega$, we explore its local neighborhood by applying Gaussian perturbations with magnitude $\sigma$ as:
\begin{align}
\label{eq:es_perturb}
\hat{\tau}_i
=
\tau_i+ \epsilon_i,
\quad
\epsilon_i \sim \mathcal{N}(0,\sigma^2I).
\end{align}
Following the principle of ES, the evolution objective is to maximize the execution feedback of perturbed action trajectories. Mathematically, for any trajectory $\tau$, we define its expected feedback under stochastic perturbations as:
\begin{equation}
\begin{aligned}
\mathcal{J}(\tau)
&= \mathbb{E}_{\epsilon}[J(\tau+\epsilon)] \\
&= \int J(\tau+\epsilon)\cdot p(\epsilon)\,d\epsilon .
\end{aligned}
\end{equation}
Therefore, the gradient of $\mathcal{J}(\tau)$ with respect to $\tau$ provides the steepest ascent direction as:
\begin{align}
\label{eq:es_derive1}
\nabla_{\tau}\mathcal{J}(\tau)
&=
\nabla_{\tau}
\int
J(\tau+\epsilon)\cdot p(\epsilon)\,d\epsilon
\nonumber\\
&=
\int
J(x)\cdot\nabla_{\tau}p(x-\tau)\,dx
\quad\quad (x=\tau+\epsilon)
\nonumber\\
&=
\int
J(x)\cdot p(x-\tau)
\cdot\nabla_{\tau}\log p(x-\tau)\,dx
\nonumber\\
&=
\mathbb{E}_{x\sim p(x-\tau)}
\left[
J(x)\cdot\nabla_{\tau}\log p(x-\tau)
\right].
\end{align}
Since the perturbation noise follows a Gaussian distribution in Eq.~\eqref{eq:es_perturb}, the derivative term in Eq.~\eqref{eq:es_derive1} is:
\begin{align}
\label{eq:es_derive2}
\nabla_{\tau}\log p(x-\tau)
&=
\nabla_{\tau}
\log
\left(
\frac{1}{\sqrt{2\pi}\sigma}
\exp
\left(
-\frac{\|x-\tau\|^2}{2\sigma^2}
\right)
\right)
\nonumber\\
&=
\nabla_{\tau}
\left(
-\frac{\|x-\tau\|^2}{2\sigma^2}
+C
\right)
\nonumber\\
&=
\frac{x-\tau}{\sigma^2}
=
\frac{\epsilon}{\sigma^2},
\end{align}
Note that the expectation over $x\sim p(x-\tau)$ in Eq.~\eqref{eq:es_derive1} is equivalent to the expectation over the perturbation noise $\epsilon$. Consequently, substituting Eq.~\eqref{eq:es_derive2} into Eq.~\eqref{eq:es_derive1} yields:
\begin{align}
\label{eq:direction}
\nabla_{\tau}\mathcal{J}(\tau)
=
\frac{1}{\sigma^2}
\mathbb{E}_{\epsilon}
\left[
J(\tau+\epsilon)\epsilon
\right].
\end{align}
For each action trajectory, this result demonstrates that the feedback-weighted perturbations provide an unbiased estimate of the evolution direction. Then, the overall evolution direction of the action trajectory distribution is given as:
\begin{align}
\label{eq:dist_direction}
\mathbb{E}_{\tau\sim p_{\theta}(\tau\mid\mathcal{S})}
\left[
\nabla_{\tau}\mathcal{J}(\tau)
\right]
&=
\mathbb{E}_{\tau\sim p_{\theta}(\tau\mid\mathcal{S})}
\left[
\frac{1}{\sigma^2}
\mathbb{E}_{\epsilon}
\left[
J(\tau+\epsilon)\epsilon
\right]
\right] \nonumber\\
&=
\frac{1}{\sigma^2}
\mathbb{E}_{\tau\sim p_{\theta}(\tau\mid\mathcal{S}),\,\epsilon}
\left[
J(\tau+\epsilon)\epsilon
\right].
\end{align}

\subsection{Self-Supervision for Parameter Adaptation}
Although Eq.~\eqref{eq:dist_direction} provides the evolution direction in the trajectory space, it must be transferred to the parameter space for online model adaptation. To bridge the two spaces, we construct a self-supervised MSE loss based on the sampled action trajectories in Eq.~\eqref{eq:t_sample} and their perturbed variants in Eq.~\eqref{eq:es_perturb}, weighted by the execution feedback as:
\begin{align}
\label{eq:mse}
\mathcal{L} = \frac{1}{|\Omega|}\sum_{\tau_i \in \Omega}J(\hat{\tau}_i) \cdot \| \hat{\tau}_i - \tau_i \|_2^2 ,
\end{align}
where $J(\hat{\tau}_i)$ denotes the execution feedback of $\hat{\tau}_i$. Although $\hat{\tau}_i$ is generated from $\tau_i$, we detach $\hat{\tau}_i$ from the computation graph during implementation. Consequently, the gradients are back-propagated only through $\tau_i$ to update the model parameters. The resulting gradient of $\mathcal{L}$ with respect to the model parameters $\theta$ is given by:
\begin{align}
\label{eq:loss_gradient}
\nabla_{\theta}\mathcal{L}
&=\frac{1}{|\Omega|}\sum_{\tau_i\in\Omega}J(\hat{\tau}_i) \cdot \nabla_{\theta}\|\hat{\tau}_i-\tau_i\|_2^2
\nonumber\\
&=-\frac{2}{|\Omega|}\sum_{\tau_i \in \Omega}J(\hat{\tau}_i) \cdot (\hat{\tau}_i-\tau_i) \cdot \nabla_{\theta}\tau_i
\nonumber\\
&=-\frac{2}{|\Omega|}\sum_{\tau_i \in \Omega}J(\hat{\tau}_i) \cdot \epsilon_i \cdot \nabla_{\theta}\tau_i ,
\end{align}
where $\nabla_{\theta}\tau_i=\nabla_{\theta}\,\pi_\theta(\mathcal{S},\boldsymbol{z}_i)$. For each action trajectory $\tau$, taking expectation over its perturbation noise $\epsilon$ gives:
\begin{align}
\mathbb{E}\left[\nabla_{\theta}\mathcal{L}\right]
&= -2 \cdot \nabla_{\theta}\tau \cdot \mathbb{E} \left[ J(\hat{\tau}) \cdot \epsilon \right]
\nonumber\\
&= -2 \cdot \nabla_{\theta}\tau \cdot \mathbb{E} \left[ J(\tau+\epsilon) \cdot \epsilon \right]
\nonumber\\
&\propto -\nabla_{\theta}\tau \cdot \nabla_{\tau}\mathcal{J}(\tau).
\end{align}
This result shows that the MSE loss projects the evolution direction in Eq.~\eqref{eq:direction} into the parameter space through the model Jacobian $\nabla_{\theta}\tau$. The negative sign will be canceled by the gradient descent rule used to minimize the loss. Similarly, the expectation over all samples in $\Omega$ also projects the overall evolution direction in Eq.~\eqref{eq:dist_direction} into the parameter space.

\subsection{Failure-Aware Regularization}
\begin{figure}[t]
    \centering
    \includegraphics[width=1\linewidth]{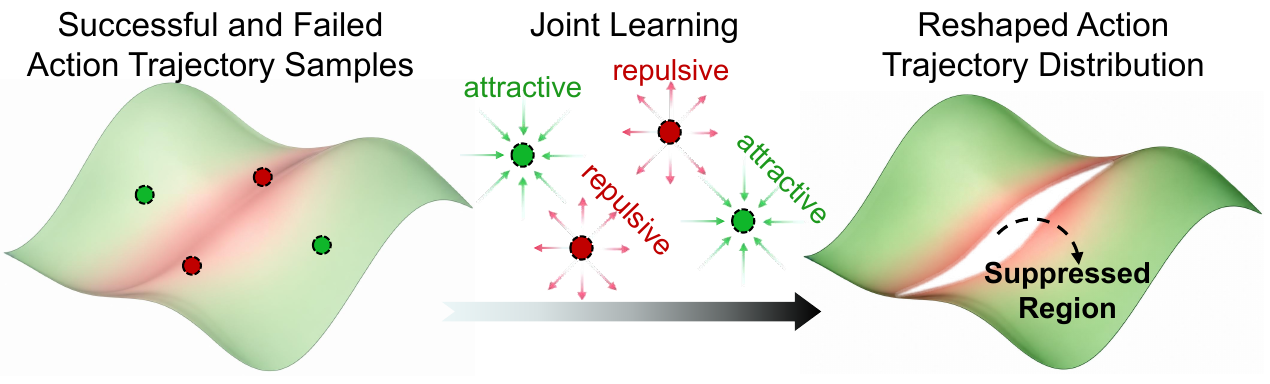}
    \vspace{-6mm}
    \caption{Illustration of how successful and failed experiences jointly reshape the action trajectory distribution.}
    \label{fig:reshape}
\end{figure}

Although successful action trajectories provide positive guidance for model adaptation, failed trajectories also contain valuable information that has not yet been exploited. We follow the same intuition as navigating a maze: once a route proves to be a dead end, it should not be revisited. Likewise, a robot should learn to avoid generating action trajectories that have been proven to fail. To this end, we modify the execution feedback in Eq.~\eqref{eq:reward} as:
\begin{align}
J(\tau)=
\begin{cases}
1, & \text{if the task succeeds by executing } \tau,\\
-1, & \text{otherwise}.
\end{cases}
\label{eq:sfscore}
\end{align}
Thus, the gradient of $\mathcal{L}$ in Eq.~\eqref{eq:loss_gradient} can be regularized as:
\begin{align}
\label{eq:sf_gradient}
\nabla_{\theta}\mathcal{L}
&=
-\frac{2}{|\Omega|} \left(
\sum_{\tau_i\in\Omega_S}
\epsilon_i\nabla_{\theta}\tau_i
-
\sum_{\tau_j\in\Omega_F}
\epsilon_j\nabla_{\theta}\tau_j \right),
\end{align} 
where $\Omega = \Omega_S \cup \Omega_F$, with $\Omega_S=\{\tau_i\in\Omega|J(\tau_i)=1\}$ and $\Omega_F=\{\tau_i\in\Omega|J(\tau_i)=-1\}$ denoting the subsets of successful and failed trajectories, respectively. As illustrated in Fig.~\ref{fig:reshape}, the joint effect of the two gradient terms can be interpreted as follows: the successful term provides an attractive signal that pulls the trajectory distribution toward high-feedback regions, while the failure term provides a complementary repulsive signal that pushes the distribution away from previously explored failure regions.

\section{Experiment}
\subsection{Implementation Details}
\textbf{Online Adaptation Procedure. }Algorithm~\ref{alg:framework} summarizes the proposed online adaptation framework for a single task scenario. It can be extended to multi-task scenarios by parallelizing across multiple GPUs, where each GPU collects action trajectories for one task and synchronizes multi-GPU gradients before policy updates. In real-world robot execution, the perturbed action trajectories may induce joint oscillations due to high-frequency variations, which can occur even in the original policy outputs. Therefore, we apply mathematical tools, such as quintic polynomial fitting or spline interpolation, to improve the smoothness of action trajectories before execution. This smoothing constraint does not affect the derivation in Sec.~\ref{sec:evolution_direction}, since $J(\hat{\tau})$ serves as an independent weighting term in the gradient estimation. Moreover, we control the ratio between the successful and failed trajectories in Eq.~\eqref{eq:sf_gradient}, with $|\Omega_S| > 2|\Omega_F|$ and \mbox{$|\Omega| =10$}. Empirically, this constraint allows the positive evolution direction to dominate the gradient estimation, leading to a more stable online adaptation process.

\begin{algorithm}[t]
\small
\caption{Online-ES Adaptation Framework}
\label{alg:framework}
\KwIn{Interactive Environment $Env$, \\
      \qquad\,\,\,\, Pretrained Flow-Matching VLA policy $\pi_\theta$\;}

Initialize rollout buffer $\Omega$\;

\While{online adaptation is not terminated}{

    Reset $Env$ to the same initial state\;
    Initialize action trajectory buffer ($\tau$, $\hat{\tau}$)\;
    \While{the task is not terminated in $Env$}{
        Get observations from $Env$\;
        Generate a random latent noise vector\;
        Generate an action chunk using Eq.~\eqref{eq:ode}\;
        Perturb the action chunk using Eq.~\eqref{eq:es_perturb}\;
        Execute the perturbed action chunk in $Env$\;
        Append the raw\,/\,perturbed action chunks to ($\tau$,\,$\hat{\tau}$)\;
    }

    Append ($\tau$, $\hat{\tau}$) with its execution feedback $J(\hat{\tau})$ to $\Omega$\;

    \If{enough trajectories are collected in $\Omega$}{
        Compute the self-supervised loss using Eq.~\eqref{eq:mse}\;
        Update $\pi_\theta$ using the gradient in Eq.~\eqref{eq:sf_gradient}\;
        Clear rollout buffer $\Omega$\;
    }

}
\end{algorithm}

\textbf{Baseline Model.} We adopt GR00T N1.5 \cite{bjorck2025gr00t} as the baseline model, augmented with depth-based 3D features following GeoVLA \cite{sun2026geovla}. This baseline model consists of a vision encoder, a depth encoder, a language encoder, and a DiT-based action expert. The features extracted from the first three modules serve as conditional inputs of the action expert for action generation. During the online adaptation stage, we only finetune the parameters of the action expert, while keeping other modules frozen.

\textbf{Hardware Resources.} In simulation experiments, we utilize 8 H100 GPUs for parallel model adaptation, with each GPU responsible for one task scenario. For real-world experiments, we build the task scenario with an AgiBot G1 robot \cite{ZhiyuanG1}, which consists of dual arms with 14-DoF joints and two parallel-jaw grippers. Since only one task scenario is involved in the real-world setting, the model adaptation uses only a single H100 GPU.

\subsection{Simulation Results}
We conducted simulation experiments on the LIBERO benchmark \cite{liu2023libero}, which consists of 4 task suites with 10 tasks per suite. We first fine-tune the baseline model on the official human demonstration dataset using offline supervised learning (Offline-SFT) for 30000 training steps. The finetuned model then interacts with the benchmark environments and is continuously refined using our proposed Online-ES. 


\begin{figure}[t]
    \centering
    \includegraphics[width=1\linewidth]{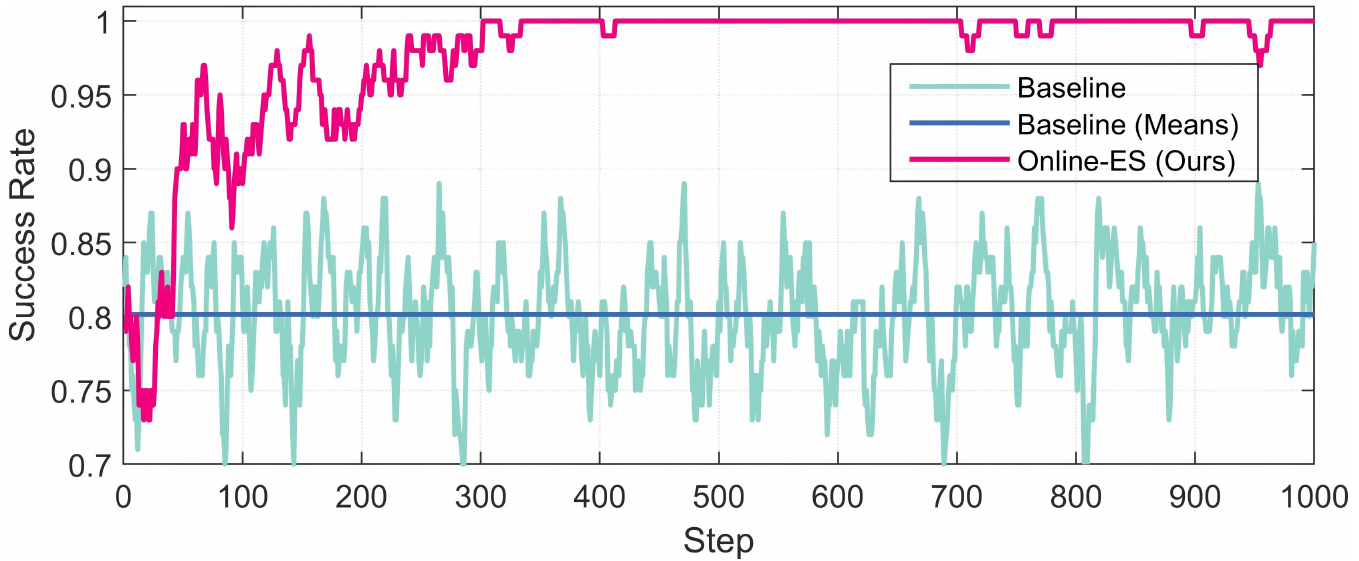}
    \vspace{-6mm}
    \caption{The curve of single-task adaptation with fixed scene initialization.}
    \label{fig:stsp}
\end{figure}

\textbf{Single-Task Adaptation with Fixed Scene Initialization.} In this setting, we randomly select one task from the LIBERO benchmark and perform online adaptation under a fixed scene configuration (i.e., same object placement). The perturbation magnitude $\sigma$ in Eq.~\eqref{eq:es_perturb} is set to 0.03. Fig.~\ref{fig:stsp} shows the success rate curves on the 7$_{th}$ task of the LIBERO Spatial benchmark. For a clear comparison, we also use the Offline-SFT baseline to generate action trajectories and compute the success rates synchronously. As we can see, Online-ES framework improves the average success rate to nearly 100\% after 300 adaptation steps.

\begin{figure}[t]
    \centering
    \includegraphics[width=1\linewidth]{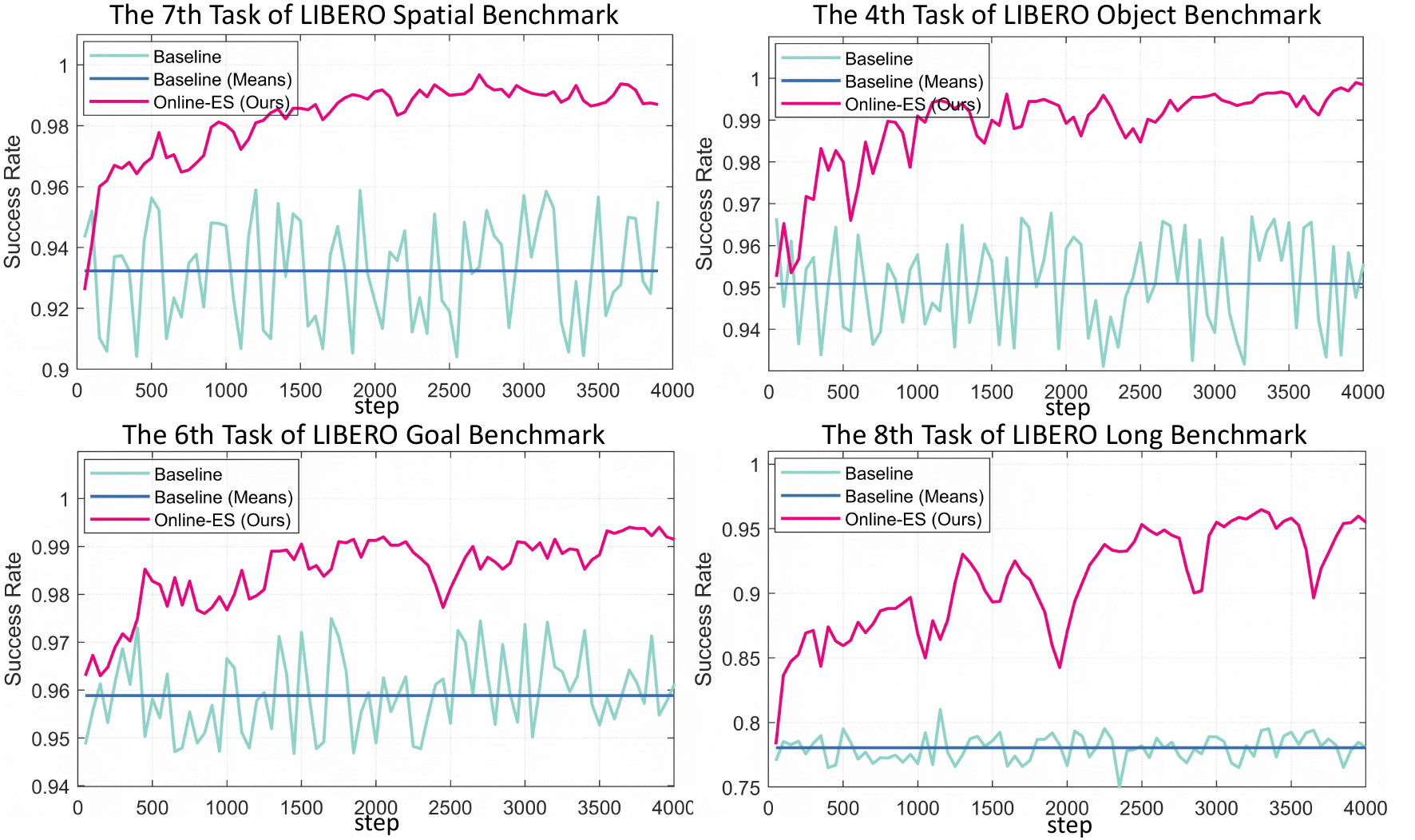}
    \vspace{-6mm}
    \caption{The curves of single-task adaptation with random scene initialization.}
    \label{fig:stmp}
\end{figure}

\begin{table}[t]
\centering
\caption{The results of single-task adaptation with random scene initialization on  LIBERO benchmark.}
\label{tab:stmp}
\begin{tabular}{c|cccc}
\hline
 Spatial           &  $1st$ Task          & $7th$ Task           & $8th$ Task           & $9th$ Task \\ 
Baseline       & 0.922 \;\;             & 0.934 \;\;             & 0.916 \;\;             & 0.902 \;\;             \\ 
Online-ES  & 0.994 $\uparrow$              & 0.992 $\uparrow$             & 0.996 $\uparrow$              & 0.970 $\uparrow$             \\ \hline
 Object            &          $3rd$ Task  &          $4th$ Task  &          $7th$ Task  & $9th$ Task  \\ 
Baseline       & 0.976 \;\;             & 0.952 \;\;             & 0.980 \;\;             & 0.974 \;\;             \\ 
Online-ES & 0.994 $\uparrow$             & 0.996 $\uparrow$              & 0.996 $\uparrow$             & 0.998 $\uparrow$             \\ \hline
 Goal              &  $3rd$ Task          & $5th$ Task           & $6th$ Task           & $9th$ Task    \\ 
Baseline       & 0.912 \;\;             & 0.916 \;\;             & 0.960 \;\;             & 0.934 \;\;             \\ 
Online-ES & 0.960 $\uparrow$             & 0.988 $\uparrow$             & 0.992 $\uparrow$             & 0.978 $\uparrow$             \\ \hline
 Long             &  $4th$ Task    &  $6th$ Task    & $8th$ Task    & $9th$ Task    \\
Baseline       &  0.912 \;\;            & 0.860 \;\;                  &  0.784 \;\;                 & 0.902 \;\;                  \\ 
Online-ES &  0.970 $\uparrow$            & 0.984 $\uparrow$                  &  0.956 $\uparrow$                 & 0.968 $\uparrow$                  \\ \hline
\end{tabular}
\end{table}

\textbf{Single-Task Adaptation with Random Scene Initialization.} In this setting, we randomly select one task from the LIBERO benchmark and perform adaptation across different scene configurations (i.e., 50 object placements). The perturbation magnitude $\sigma$ is set to 0.03. Fig.~\ref{fig:stmp} shows the success-rate curves for one representative task from each of the four task suites. Although the success rate consistently improves for all tasks, it increases at a slower pace compared with Fig.~\ref{fig:stsp}, and the final performance remains below 100\%. This degradation may result from the variations in visual observations caused by different object placements, which can lead to conflicting gradients. To further validate the effectiveness, we have selected more tasks to perform online adaptation. Table~\ref{tab:stmp} reports the success rates.

\begin{figure}[t]
    \centering
    \includegraphics[width=1\linewidth]{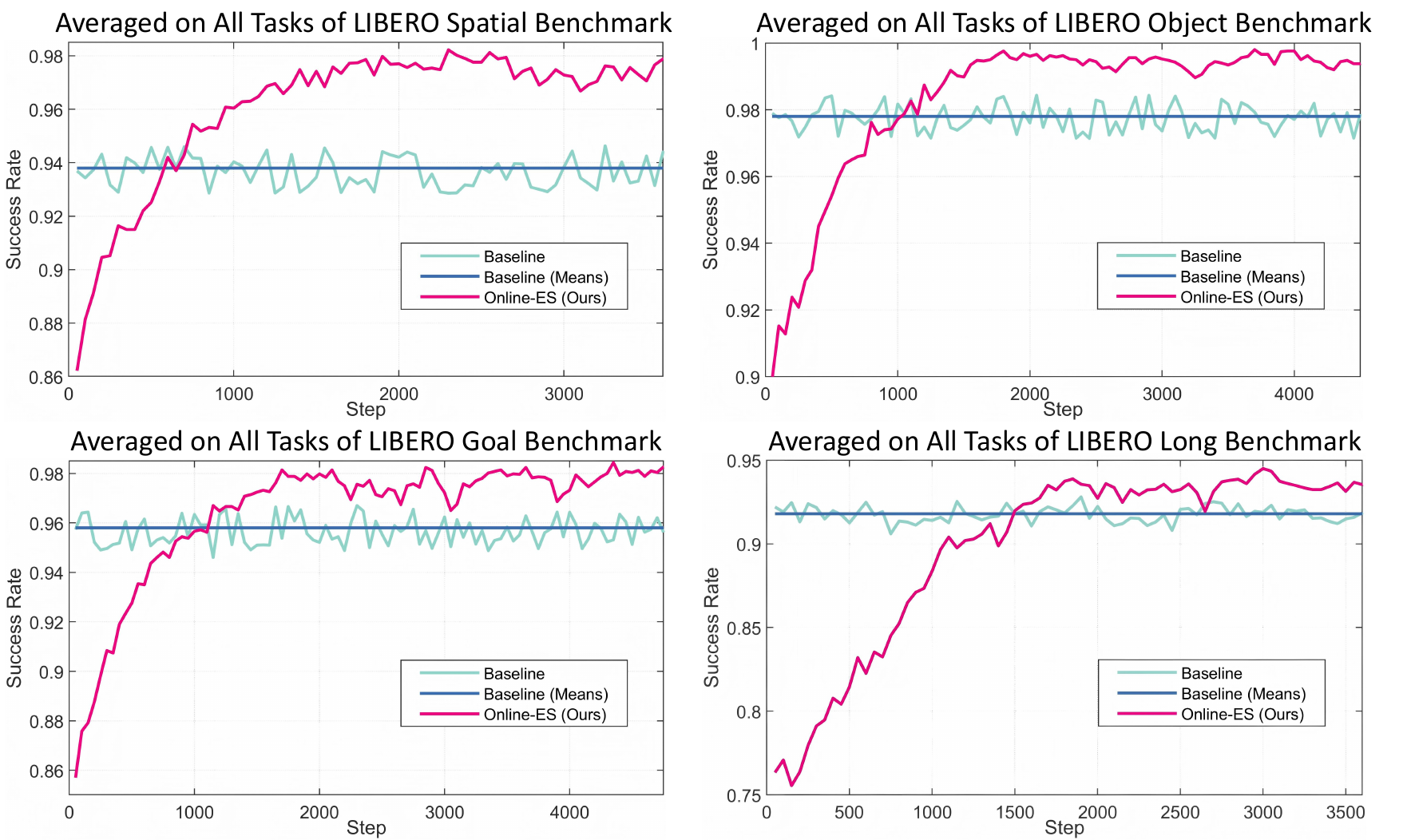}
    \vspace{-6mm}
    \caption{The curves of multi-task adaptation with random scene initialization.}
    \label{fig:mtmp}
\end{figure}

\begin{table}[t]
\centering
\caption{The results of multi-task adaptation with random scene initialization on LIBERO benchmark. * represents the best.}
\label{tab:mtmp}
\begin{tabular}{c|ccccc}
\hline
                  & Spatial        & Object        & Goal        & Long    & Average \\ \hline
\textscale{0.8}{OpenVLA\cite{openvla}} & 0.847    & 0.884        & 0.792       & 0.537      & 0.765 \\
\textscale{0.8}{VLA-RL\cite{lu2025vla}} & 0.902   & 0.918        & 0.822       & 0.598      & 0.810 \\
\textscale{0.8}{VLA-RFT \cite{li2025vla}} & 0.944  & 0.944       & 0.954       & 0.802      & 0.911  \\
$\pi_{0.5}$ \cite{pmlr-v305-black25a} & 0.988     & 0.982    &  0.980  &   0.924 &  0.969  \\
$\pi_{RL}$\cite{chen2025pirl}   & 0.996*  & 1.000* & 0.996* & 0.940* & 0.983*  \\ \hline
Baseline       & 0.938 \;\;         & 0.978 \;\;        & 0.958 \;\;      & 0.918 \;\;  & 0.948 \;\; \\ 
Online-ES      & 0.978 $\uparrow$         & 0.996 $\uparrow$         & 0.980 $\uparrow$      & 0.936 $\uparrow$   & 0.972 $\uparrow$  \\ \hline
\end{tabular}
\end{table}

\begin{figure}[t]
    \centering
    \includegraphics[width=1\linewidth]{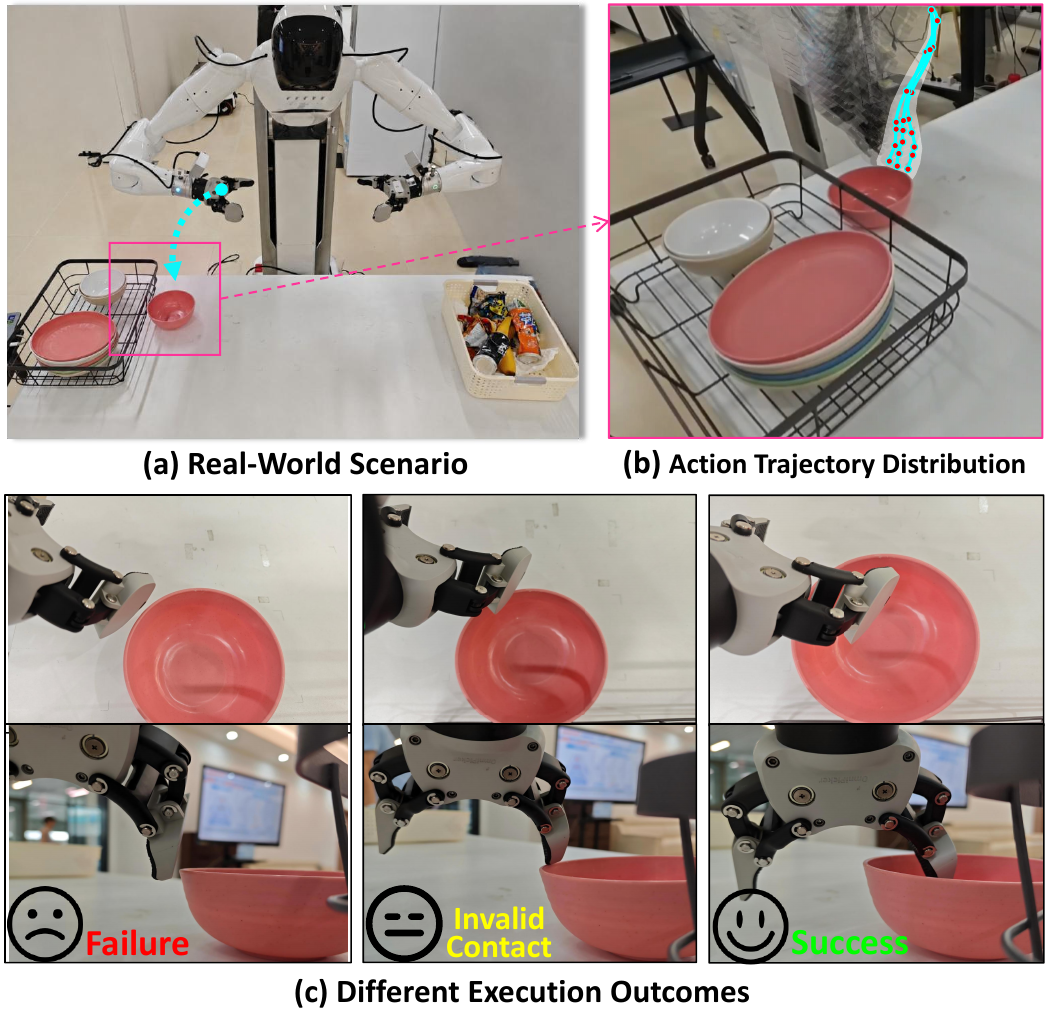}
    \vspace{-6mm}
    \caption{Real-world experimental setup. (a) The task scenario in which the robot attempts to grasp the pink bowl. (b) The action trajectory distribution visualized by overlaying the arm motions (see the ghosted region) from multiple rollouts. (c) Three execution outcomes with top views shown above and side views shown below.}
    \label{fig:real}
\end{figure}

\textbf{Multi-Task Adaptation with Random Scene Initializations.} In this setting, we perform joint online adaptation across all tasks within each task suite (10 tasks $\times$ 50 different object placements) of the LIBERO benchmark. During online adaptation, 8 GPUs perform rollouts in parallel, with each GPU handling one task. Accounting for the increased difficulty, we adopt an annealing strategy for the perturbation magnitude $\sigma$, linearly decreasing it from 0.3 to 0.03 over the first 2,500 steps. Fig.~\ref{fig:mtmp} shows the success rate curves for the four task suites. The larger exploration range initially leads to substantially lower success rates than the baseline, but performance gradually surpasses the baseline as the exploration range decreases. Table~\ref{tab:mtmp} reports the quantitative results, with VLA-RL based on OpenVLA and $\pi_{RL}$ based on $\pi_{0.5}$ for reinforcement fine-tuning. Despite not achieving the highest success rate, Online-ES achieves competitive policy improvement with its simpler adaptation scheme.

\begin{figure}[t]
    \centering
    \includegraphics[width=1\linewidth]{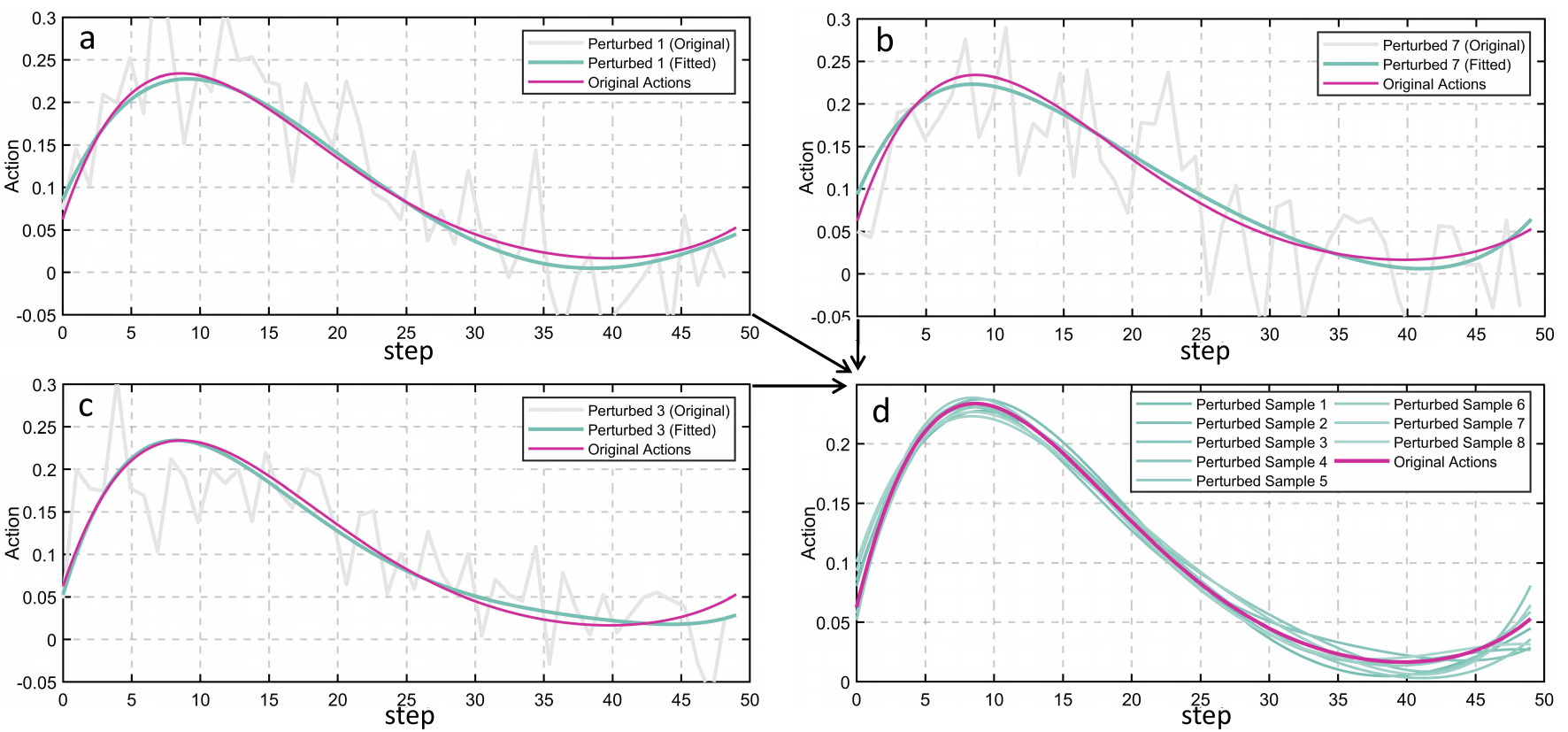}
    \vspace{-6mm}
    \caption{Real-World Action Trajectory Perturbation. (a–c) Three examples of perturbed trajectories (gray) and refitted trajectories (cyan) generated from the same original action trajectory (magenta). (d) The local exploration region around the raw action trajectory formed by multiple perturbations.}
    \label{fig:realdist}
\end{figure}

\begin{table}[t]
\centering
\caption{The Quantitative Results of Real-World Online Adaptation, which are evaluated over 100 independent trials respectively.}
\label{tab:real}
\begin{tabular}{l|ccc}
\hline
            & \textcolor{red}{Failure} & \begin{tabular}[c]{@{}c@{}}\textcolor{yellow}{Invalid}\\ \textcolor{yellow}{Contact}\end{tabular} & \textcolor{green}{Success}  \\ \hline
Baseline    & 27\,/\,100 \;\;  & 41\,/\,100 \;\;         & 32\,/\,100 \;\;            \\
Online-ES (50 steps) & 22\,/\,100 $\downarrow$  & 42\,/\,100 $\uparrow$          & 36\,/\,100 $\uparrow$           \\ 
Online-ES (100 steps) & 19\,/\,100 $\downarrow$  & 37\,/\,100 $\downarrow$          & 44\,/\,100 $\uparrow$           \\ 
Online-ES (150 steps) & 12\,/\,100 $\downarrow$  & 34\,/\,100 $\downarrow$          & 54\,/\,100 $\uparrow$           \\ \hline
\end{tabular}
\end{table}

\subsection{Real-World Results}
Fig.~\ref{fig:real} (a) shows the real-world experimental setup for the tabletop clearing task. For ease of environment resetting, we select the subtask of \textbf{\textit{grasping the pink bowl}} as the evaluation benchmark. We collect 300 human teleoperation demonstrations and train the policy offline for 30,000 steps to obtain a baseline checkpoint. By projecting action trajectories generated from multiple rollouts onto a single image, we can observe a coarse outline of the underlying action trajectory distribution as shown in Fig.~\ref{fig:real} (b). To evaluate the model performance, we categorize the grasping outcomes into three classes, as illustrated in Fig.~\ref{fig:real} (c). \textbf{\textit{Failure}} denotes cases where the gripper completely misses the bowl. \textbf{\textit{Invalid Contact}} refers to cases where the gripper reaches the bowl rim with an improper grasping pose. Although these cases also fail to complete the task, they are considered closer to successful grasping than \textbf{\textit{Failure}}. Finally, \textbf{\textit{Success}} denotes the cases where the gripper has a proper grasping pose, with the bowl rim positioned between the two fingers.

In real-world experiments, robot motors are highly sensitive to high-frequency input variations, and perturbing the action trajectory according to Eq.~\eqref{eq:es_perturb} can induce severe joint oscillations. We follow the post-processing strategy \cite{zhao2025rail} to refit the perturbed action trajectories using a high-order polynomial and then execute them. As illustrated in Fig.~\ref{fig:realdist}, this procedure transforms the independent noise added to individual actions into a smooth trajectory-level perturbation over a longer horizon, thereby still forming a local exploration region around the original action trajectory.

We adapted the model online for 150 steps. The perturbation magnitude $\sigma$ is set to 0.05 for the AgiBot G1 robot. At each step, we fill the buffer $\Omega$ with 10 action trajectories and compute their averaged gradients to update the model parameters. Note that the ratio of successful to failed trajectories in $\Omega$ is maintained at 8:2. When the baseline checkpoint performs poorly, we have to run more than 10 trials to collect action trajectories. Tab.~\ref{tab:real} reports the quantitative results, where each row is evaluated over 100 rollouts. We can observe clear decreasing trends for both \textbf{\textit{Failure}} cases and \textbf{\textit{Invalid Contact}} cases, demonstrating that the action trajectory distribution of the policy has been progressively refined.

\begin{figure}[t]
    \centering
    \includegraphics[width=1\linewidth]{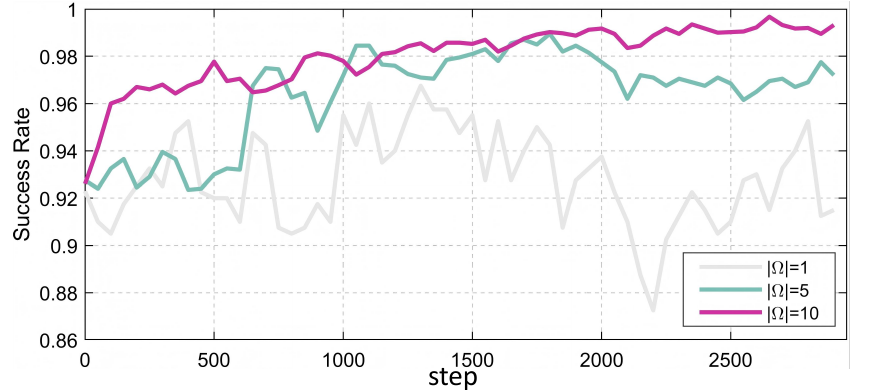}
    \vspace{-3mm}
    \caption{The success rate curves with different buffer sizes $\lvert\Omega\rvert$.}
    \label{fig:ablation_omega}

    \vspace{4mm} 
    \includegraphics[width=1\linewidth]{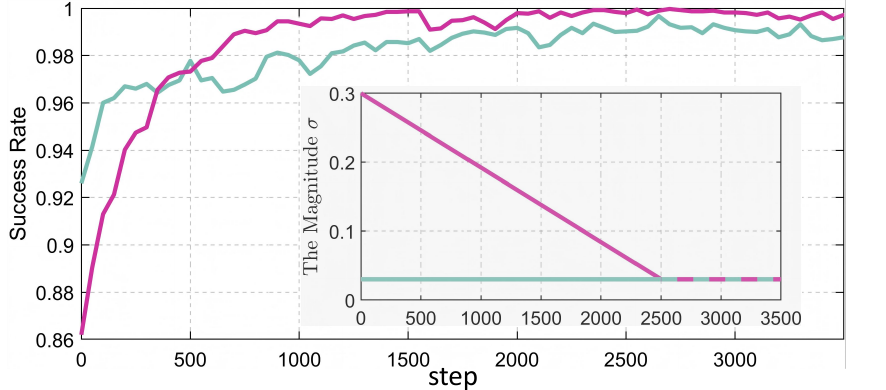}
    \vspace{-3mm}
    \caption{The success rate curves with different perturbation magnitudes $\sigma$.}
    \label{fig:ablation_std}
\end{figure}

\subsection{Ablation Studies}
\textbf{The size of action trajectory buffer $\Omega$ in Eq.~\eqref{eq:t_sample}.} This hyperparameter controls how well each online-adaptation step captures the current action trajectory distribution, with a larger $\lvert\Omega\rvert$ providing a more comprehensive estimate. We conducted this ablation study on the 7th task of the LIBERO Spatial benchmark. Fig.~\ref{fig:ablation_omega} shows the success rate curves for $\lvert\Omega\rvert$ set to 1, 5, and 10. As $\lvert\Omega\rvert$ increases, the success rate exhibits a smoother and more stable upward trend. 

\textbf{The effect of perturbation magnitude $\sigma$ in Eq.~\eqref{eq:es_perturb}.} This hyperparameter controls the exploration range, with $\sigma$ specifying the standard deviation of the injected noise. We again conducted this ablation study on the 7th task of the LIBERO Spatial benchmark. Fig.~\ref{fig:ablation_std} shows the success rate curves for a fixed $\sigma=0.03$ and a linear annealing schedule that decays $\sigma$ from 0.3 to 0.03 over 2500 adaptation steps. It can be observed that both curves exhibit a clear upward trend. In the early stage, the annealed setting achieves substantially lower success rates due to its larger exploration range. As $\sigma$ decreases, its performance catches up with and eventually surpasses that of the fixed-$\sigma$ setting. This phenomenon suggests that appropriately broad exploration can capture valuable information that may be missed by a narrow exploration range.

\begin{figure}[t]
    \centering
    \includegraphics[width=1\linewidth]{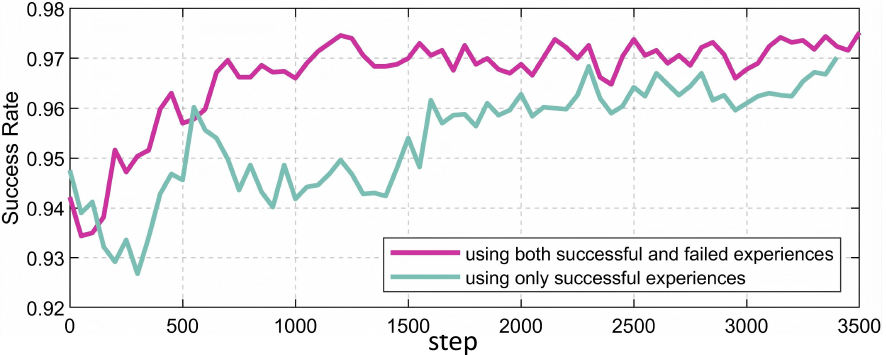}
    \vspace{-6mm}
    \caption{The success rate curves with a hybrid buffer of successful\,/\,failed trajectories versus a success-only buffer.}
    \label{fig:ablation_sf}
\end{figure}

\textbf{The effect of failure experiences in Eq.~\eqref{eq:sf_gradient}.} The gradients derived from failure experiences can regularize the evolution direction, discouraging the model from revisiting action trajectories that have previously led to failure. To verify this effect, we conducted this ablation study across all tasks of the LIBERO Spatial benchmark. Fig.~\ref{fig:ablation_sf} shows the average success rate curves for online adaptation with successful experiences only and mixed experiences. The results demonstrate that failure-based regularization can accelerate the convergence of success rate.

\section{CONCLUSION}

In this paper, we propose an Evolution Strategy-based online adaptation framework for Flow Matching VLAs, which refines the learned action trajectory distribution through execution feedback. Our key observation is that task failures may arise from an ill-formed trajectory distribution rather than the lack of feasible solutions. By exploiting the stochasticity of Flow Matching, our method performs trajectory-space exploration and adapts model parameters through a self-supervised objective with failure-aware regularization. Extensive experiments in simulation and real-world robotic manipulation tasks demonstrate that our framework effectively improves the performance of Flow Matching VLAs without additional human demonstrations. Future work will focus on improving the efficiency of utilizing the interaction experiences for online adaptation.

\bibliographystyle{IEEEtran}
\bibliography{reference}

\end{document}